\documentclass[conference]{IEEEtran}
\IEEEoverridecommandlockouts
\usepackage{cite}
\usepackage{amsmath,amssymb,amsfonts}
\usepackage{algorithmic}
\usepackage{graphicx}
\usepackage{textcomp}
\usepackage{xcolor}
\def\BibTeX{{\rm B\kern-.05em{\sc i\kern-.025em b}\kern-.08em
    T\kern-.1667em\lower.7ex\hbox{E}\kern-.125emX}}

\newtheorem{theorem}{Theorem}

\newtheorem{proof}{Proof}

\usepackage{multicol, multirow, subfigure}
\usepackage{makecell, float}
\usepackage{booktabs}
\usepackage{cite}
\makeatletter

\newcommand{\Rmnum}[1]{\expandafter\@slowromancap\romannumeral #1@}
\makeatother

\begin{document}

\title{Domain Generalization via Discrete Codebook Learning}

\author{Shaocong Long$^{1\star}$ \quad Qianyu Zhou$^{2\star}$ \quad Xikun Jiang$^3$ \quad Chenhao Ying$^{1\dag}$ \quad Lizhuang Ma$^1$ \quad Yuan Luo$^{1\dag}$ \\
\textit{$^1$ Shanghai Jiao Tong University, Shanghai, China } \\
\textit{$^2$ Jilin University, Changchun, China} \\
\textit{$^3$ Copenhagen University, Copenhagen, Denmark}\\
\thanks{This work was supported in part by National Key R\&D Program of
China under Grant 2022YFA1005000. $^\star$Equal contributions. $^\dag$Corresponding authors (yingchenhao@sjtu.edu.cn, yuanluo@sjtu.edu.cn). }
}

\maketitle

\begin{abstract}
Domain generalization (DG) strives to address distribution shifts across diverse environments to enhance model's generalizability. Current DG approaches are confined to acquiring robust representations with continuous features, specifically training at the pixel level. However, this DG paradigm may struggle to mitigate distribution gaps in dealing with a large space of continuous features, rendering it susceptible to pixel details that exhibit spurious correlations or noise. In this paper, 
we first theoretically demonstrate that the domain gaps in continuous representation learning can be reduced by the discretization process. Based on this inspiring finding, 
we introduce a novel learning paradigm for DG, termed Discrete Domain Generalization (DDG). DDG proposes to use a codebook to quantize the feature map into discrete codewords, aligning semantic-equivalent information in a shared discrete representation space that prioritizes semantic-level information over pixel-level intricacies. By learning at the semantic level, DDG diminishes the number of latent features, optimizing the utilization of the representation space and alleviating the risks associated with the wide-ranging space of continuous features. Extensive experiments across widely employed benchmarks in DG demonstrate DDG's superior performance compared to state-of-the-art approaches, underscoring its potential to reduce the distribution gaps and enhance the model's generalizability.
\end{abstract}


\section{Introduction}
\begin{figure*}[t]
    \centering
    \includegraphics[width=0.98\linewidth]{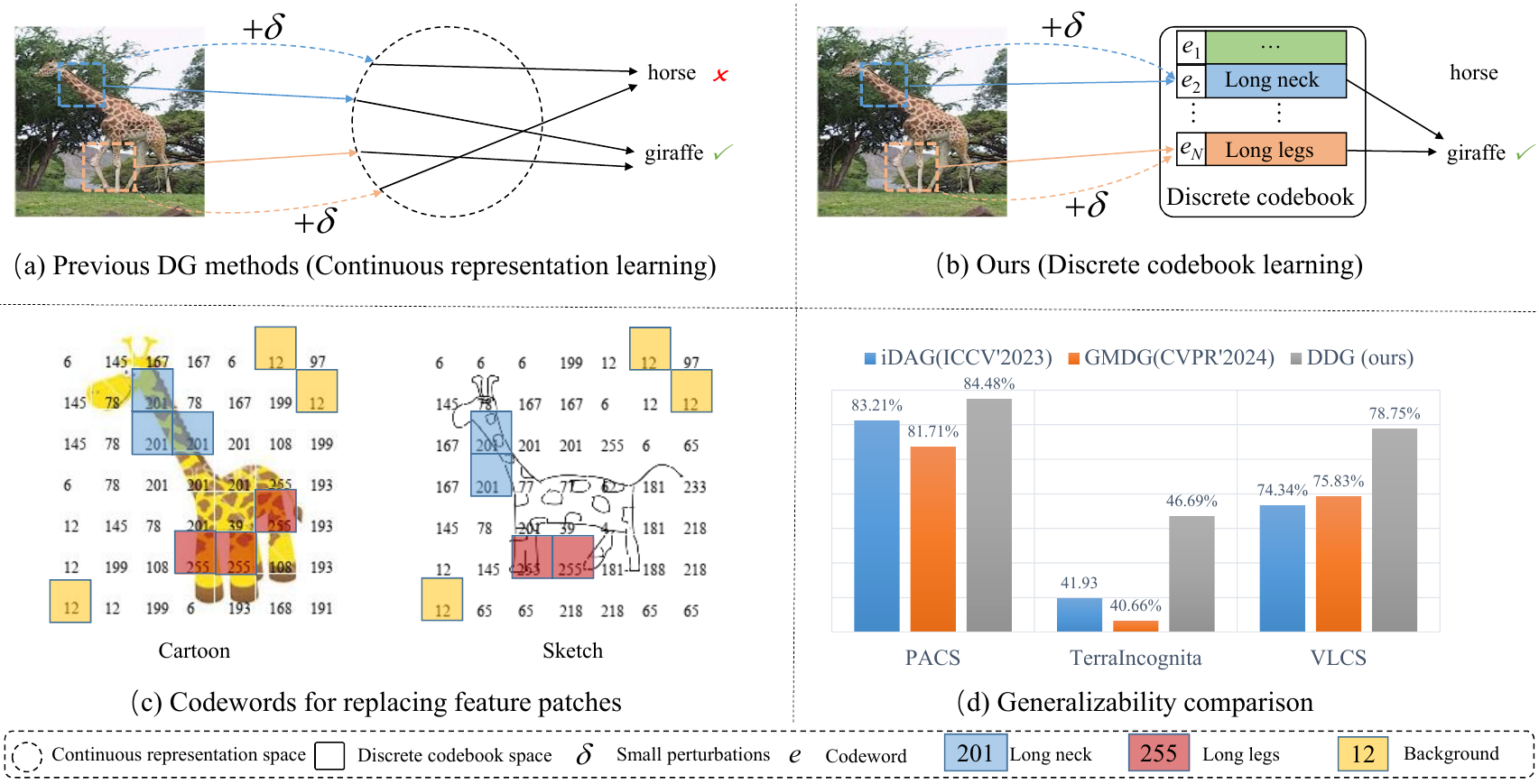}
    \vspace{-3mm}
    \caption{(a)~Existing DG methods rely on continuous representation learning, struggling with domain gaps due to large feature spaces, pixel perturbations, and interpretation. (b)~We introduce a discrete representation codebook to map features into discrete codewords, prioritizing semantic information over imperceptible pixel details, aiding distribution alignment across domains. (c)~The discretization of continuous features in our DDG. The numbers in image patches denote codeword indices. Key semantic patches in images from diverse domains~(`Cartoon' and `Sketch') are replaced with the same codeword (\emph{e.g.}, codeword 201 for \textit{long neck}, codeword 255 for \textit{long legs}). (d)~Compared to state-of-the-art DG methods, our DDG significantly improves the model's generalizability.} 
    \label{contrast}
    \vspace{-5mm}
\end{figure*}

\label{sec:intro}

Domain generalization~(DG) has garnered significant attention recently, aiming to alleviate the adverse effects of distribution shifts. DG focuses on leveraging data solely from source domains to capture essential semantic information across domains, and thus enhancing model's generalizability in target domains.
Existing DG methods have focused primarily on attaining robust features through continuous representation learning where features are represented in continuous vector space, \emph{i.e.}, learning at the pixel level, such as domain alignment~\cite{ganin2016domainadversarial,long2024rethinking, zhao2020domain}, data augmentation~\cite{zhou2024mixstyle, zhao2024style, xu2021fourier,yang2024pointdgmamba,zhou2023instance}, disentanglement~\cite{wang2022domain, zhang2022principled}, contrastive learning~\cite{yao2022pcl,kim2021selfreg}, flatness-aware strategy~\cite{cha2021swad,wang2023sharpness}, and mixture-of-experts learning~\cite{li2023sparse,zhou2022adaptive}, test-time feature shifting~\cite{jiang2024dgpic,zhou2024test}.

Despite notable progress in acquiring robust representations for DG, these methods struggle to handle distribution shifts due to the vast space of continuous features inherent in various scenarios, particularly when pixel-level correlations or noise add complexities. Depicted in Fig.~\ref{contrast}(a), each vector in the continuous representation space derived by neural networks corresponds to specific input data. This illustrates the expansive nature of continuous feature space and highlights the ability of this expressive space to capture nuanced details at the pixel level, even within semantically similar data. This presents two risks in learning continuous representations for DG: (1) Minor input perturbations can cause large feature variations, distorting semantics, and hindering accurate predictions, ultimately reducing the model's generalizability. (2) Aligning distributions in the expansive representation space induced by pixel perturbations or intricate pixel details is challenging for models with limited parameters. Furthermore, interpreting continuous representations is difficult as multiple features may map to the same semantic, complicating the interpretation.

As an effective paradigm implied by language, discrete representation learning, which encodes information in discrete codewords, has demonstrated its efficacy in generation tasks~\cite{van2017neural, esser2021taming}. Moreover, the language modality has been proven to improve the performance of vision tasks~\cite{radford2021learning}, where diverse images with the same semantics could be effectively described through a consistent text. 
%
%
This observation suggests that discrete representation learning may be inherently well-suited for various modalities that display diverse distributions. Furthermore, the use of discrete representation for inference and prediction emerges as a compelling choice, as illustrated by Fig.~\ref{contrast}(b), where, for instance, an animal exhibiting a ``long neck'' and ``long legs'' is likely to be identified as a ``giraffe''.

Motivated by the preceding analysis, we introduce a novel DG paradigm that focuses on preserving key semantic information while minimizing focus on imperceptible pixel details. This paradigm shift aims to map semantically equivalent continuous features to the same latent variable, \emph{i.e.}, learning at the semantic level. 
By reducing the number of latent variables and retaining essential features, the new paradigm optimizes the representation space and mitigates the risks associated with continuous representation learning in DG. It leverages the strengths of discrete representation and uses finite latent variables to promote domain alignment, even without domain labels, enhancing the model's generalizability.

In this study, 
we theoretically elucidate that distribution gaps across domains can be further reduced by mapping continuous feature representations into discrete codeword representations. Based on this finding, 
we propose an innovative approach for DG to address distribution shifts, named Discrete Domain Generalization~(DDG), which uses a discrete representation codebook to quantize feature maps extracted by the feature encoder into discrete codewords, offering a more stable way to capture semantic information compared to continuous representations that may also capture detrimental pixel details. As shown in Fig.~\ref{contrast}(c), our DDG understands the semantic information exhibiting diverse distributions with consistent codewords, \emph{e.g.}, codeword 201 for \textit{long neck} and codeword 255 for \textit{long legs}. The codebook serves as a quantized space of the continuous representation space and is end-to-end learnable alongside the feature encoder and classifier. Fig.~\ref{contrast}(d) demonstrates the strong results of our DDG compared with state-of-the-art (SOTA) DG methods.
Our contributions can be succinctly summarized as follows:
\begin{itemize}
    \item We propose a novel learning perspective for DG that shifts focus away from noise or imperceptible pixel details through discrete representation learning, which firstly unveils the potential of discrete representation for enhancing the model's generalizability.
    \item 
    We theoretically illustrate that the domain gaps of continuous representations can be reduced by discretization. Inspired by this, 
    we introduce a discrete codebook-based approach for DG, named Discrete Domain Generalization~(DDG), which quantizes feature maps into discrete codewords, emphasizing semantic over pixel details.
    \item We conduct comprehensive experiments on widely used DG benchmarks to showcase the superiority of our DDG over SOTA approaches. Besides, in-depth analyses verify the efficacy of discrete representations in mitigating distribution shifts and enhancing model's generalizability.
\end{itemize}


\begin{figure*}
    \centering
    \includegraphics[width=0.95\linewidth]{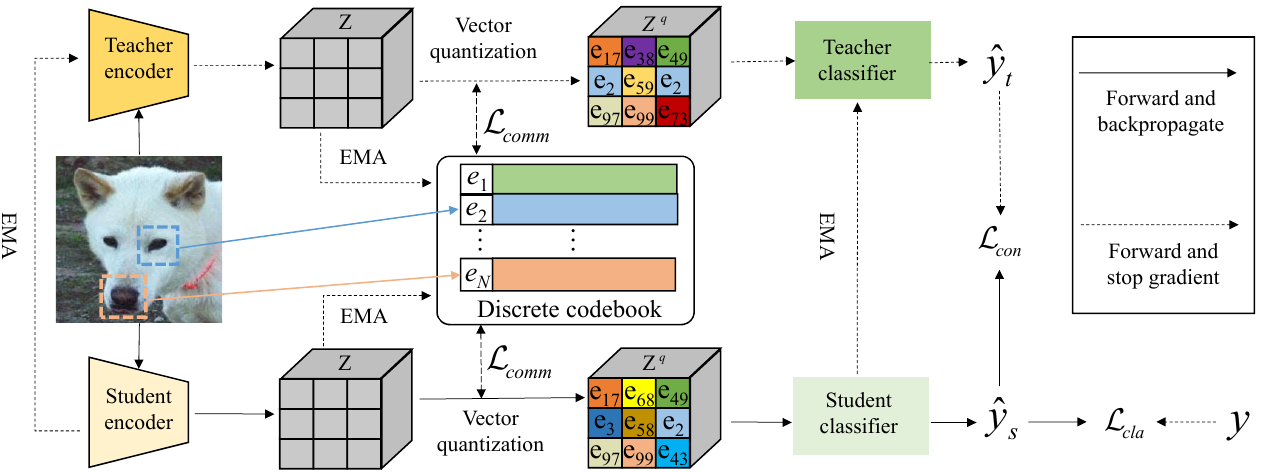}
    \vspace{-2mm}
    \caption{Framework of our Discrete Domain Generalization~(DDG). The approach uses a discrete representation codebook across domains to discretize feature maps into codewords, with predictions made by the classifier based on quantized features. The discrete codewords are chosen to replace latent variables based on their proximity. The Exponential Moving Average (EMA) of original representations is employed to optimize the codebook for heightened robustness.}
    \label{framework}
    \vspace{-3mm}
\end{figure*}


\section{Methodology}

Fig.~\ref{framework} illustrates the complete pipeline of our Discrete Domain Generalization (DDG) during the end-to-end training procedure. Our framework comprises a teacher model, a student model, and a discrete codebook. The codebook is devised to quantize the continuous features generated by the encoder into discrete codewords. In the forward computation, the quantized embedding $Z^q$, rather than the original feature map $Z$, is forwarded to the classifier. During the backward process, gradients are directly copied from the quantized feature $Z^q$ to the original feature map $Z$ in a straight-through manner. 
In the inference phase, only the student model and the codebook are retained.

    \vspace{-1mm}
\subsection{Analysis on Distribution Gaps}
    \vspace{-1mm}
Gaining a theoretical understanding of risks associated with the prevailing continuous representation learning in DG is crucial. In order to illuminate this and provide guidance for improving the generalizability, we present the following theorem from the perspective of distribution gaps across domains.

\begin{theorem}
Let $F$ denote a family of functions $f: X \rightarrow \mathbb{R}$. For two domains characterized by continuous representation distributions $P$ and $Q$ over $X$ respectively, denote the type-1 Wasserstein distance~\cite{sriperumbudur2012empirical} as $\mathcal{W}(P, Q) = \sup_{f \in F} \int |P(x)f(x) - Q(x)f(x)|dx$
. Consider a discretization function $d: X \rightarrow X_d$ that maps $x$ into the centroid of its interval, where the intervals are uniformly partitioned. Denote the discrete representation distributions as $P_d$ and $Q_d$ over $X_d$, respectively, then the following inequality holds:
\vspace{-2mm}
    \begin{equation}
        \mathcal{W}(P, Q) \geq \mathcal{W}(P_d, Q_d).
    \end{equation}
    \label{theorem1}
\end{theorem}
        \vspace{-5mm}
\begin{proof}
To facilitate the proof, consider an interval of $x\in [a, b]$, and denote $B_{\Phi}:= \sup_{f \in F} |f(x)| \geq 0$. Then the Wasserstein distance over this interval can be expressed as:
\vspace{-2mm}
    \begin{equation}
    \begin{aligned}
          \mathcal{W}_a^b(P, Q) &= \sup_{f \in F} \int_a^b |P(x)f(x) - Q(x)f(x)|dx \\
          &= B_{\Phi} \int_a^b |P(x) - Q(x)|dx \\
          & \geq B_{\Phi} |\int_a^b P(x)dx - \int_a^b Q(x)dx|. \\ \label{w_con}
    \end{aligned}
    \vspace{-5mm}
    \end{equation}
    The equality holds when $P(x) - Q(x)$ maintains the same sign for $x\in [a, b]$. This suggests that reducing distribution gaps across domains may be possible under specific constraints. However, achieving these conditions through neural network learning is challenging, as obtaining features with such distribution patterns is not straightforward.
    
    To reduce the domain gaps indicated in Eq.\eqref{w_con}, we propose adopting discrete representations. Here, $x\in[a, b]$ is discretized as $\frac{a + b}{2}$. Consequently, the discrete representation distributions in this interval are $P_d(\frac{a + b}{2}) = \int_a^b P(x)dx$, and $Q_d(\frac{a + b}{2}) = \int_a^b Q(x)dx$. The corresponding Wasserstein distance can be computed as:
    \vspace{-2mm}
    \begin{equation}
        \begin{aligned}
            \mathcal{W}_a^b(P_d, Q_d) &= \sup_{f \in F} \int_a^b |P_d(x)f(x) - Q_d(x)f(x)|dx \\
            & = f(\frac{a + b}{2}) |P_d(\frac{a + b}{2}) - Q_d(\frac{a + b}{2})| \\
            & = B_{\Phi} |\int_a^b P(x)dx - \int_a^b Q(x)dx|. \\ \label{w_dis}
        \end{aligned}
    \vspace{-5mm}
    \end{equation}
    As observed, $\mathcal{W}_a^b(P_d, Q_d)$ of our constructed discretization reaches the optimal domain gaps indicated in Eq.\eqref{w_con}.
    These principles can similarly be applied to other intervals. As a result,
    combining Eq.\eqref{w_con} and Eq.\eqref{w_dis} concludes the proof.
\end{proof}
\noindent\textbf{Remark 1.}
Theorem~\ref{theorem1} suggests that the prevalent continuous representation learning in DG may not be optimal for reducing distribution gaps to obtain robust features. This could be due to the difficulty of mitigating domain gaps when learning at the pixel level, given the vast space of continuous representations. As a result, focusing on learning semantics rather than pixels, and thereby reducing the representation space, is a promising direction to mitigate distribution shifts. To this end, we propose discretizing continuous features into discrete vectors to learn at the semantic level instead of pixel level, consequently decreasing the upper bound of the distribution gaps across domains. In this way, the upper bound of the target risk would be reduced according to the principles in~\cite{ben2010theory}, thereby promoting generalizability. These insights point to the potential of introducing discrete representation codebook learning for DG to effectively address distribution shifts.

\subsection{Discrete Representation Codebook Learning}
As indicated by Theorem~\ref{theorem1}, 
training models at the pixel level struggle with the broad representation space and may inadvertently incorporate spurious correlations or noise, compromising the model's generalizability. As a potential remedy, we attempt to curtail the number of latent variables via a discrete representation codebook, which is a promising alternative capable of mitigating the influence of redundant pixel details while preserving crucial semantic content.
To address distribution gaps between distinct domains, we advocate using the codebook to vector quantize~(VQ) features, facilitating the alignment of representations across domains. The finite codewords within the codebook can be construed as encapsulating underlying semantics common to data across domains. This codebook-driven approach diminishes the space of latent variables and thereby discards uninformative pixel-level information. Consequently, it serves as a pivotal bridge fostering alignment across domains.

In Theorem~\ref{theorem1}, the optimally reduced distribution gaps can be achieved when the features in an interval are mapped to its centroid. This finding inspires us that the discretization process for DG can be done by discretizing continuous features into the centroids of the feature exhibiting similar semantic information. Specifically, the continuous features should be mapped to the nearest vector in the discrete codebook, with both displaying similar semantic factors. Formally, we present the proposed discretization process as follows.

Denote the discrete representation codebook as $E = \{e_1, e_2, \cdots, e_N\} \in \mathbb{R}^{d_c\times N}$, where $d_c$ is the dimension of the codewords, and $N$ is the total number of the codewords. For a given feature map $Z = f(X) \in \mathbb{R}^{h \times w \times d_c}$, where $h$, $w$, and $d_c$ represent the height, width, and the number of channels, respectively, the VQ operation is applied to $Z$ to generate a discrete feature map. To ensure seamless VQ implementation, the dimension of the codewords aligns with the number of channels. The VQ operation serves to map the latent variables to discrete codewords:
\begin{equation}
   \hspace{-1mm} Z_{ij}^q = VQ(Z_{ij}) = e_k, \ \text{where}\ k = \arg\min_m||Z_{ij} - e_m||_2,
\end{equation}
where $Z^q$ denotes the quantized discrete latent variables of $Z$. In practice, the optimization of the codebook is achieved by minimizing the following objective:
\begin{equation}
\begin{aligned}
    \mathcal{L}_{disc} = \mathcal{L}_{vq} &+ \eta\mathcal{L}_{comm},\\
\hspace{-3mm}    \text{with}\ \mathcal{L}_{vq} = ||sg(Z) - Z^q||^2_2&, \ \ \mathcal{L}_{comm} = ||Z - sg(Z^q)||_2^2,
\end{aligned}
    \label{loss_vq}
\end{equation}
where $\eta$ is fixed at 0.25 for all experiments unless specified, and $sg$ signifies the stop gradient operation. The VQ loss~($\mathcal{L}_{vq}$) is used to optimize the codebook, and the commitment loss~($\mathcal{L}_{comm}$) anchors the encoder output to the codewords, thereby focusing on the semantic information and suppressing extraneous information.

In practice, we leverage the Exponential Moving Average~(EMA) of the representations to substitute the role played by the VQ loss~($\mathcal{L}_{vq}$), thereby making the codewords evolve smoothly and enhancing the robustness of the codebook. In specific, the update policy of the codebook at each iteration can be formulated as:
\begin{equation}
    \begin{split}
    N_v \leftarrow \gamma N_v + (1 - \gamma)|H|, m_v \leftarrow \gamma m_v + (1 - \gamma)\sum_{h \in H} h,
    \end{split}
\end{equation}
then the codeword is updated as $e_v = \frac{m_v}{N_v}, 1\leq v \leq N$, where $1\leq v \leq N$, and $H = \{Z_{ij}|Z^q_{ij} = e_v\}$ denotes the variable that is replaced by $e_v$. The decay factor $\gamma$ is set to 0.99, and the codebook is initialized as: $N_v = 1, m_v \sim \mathcal{N}_{d_c}(0, 1), e_v = m_v$.

Additionally, a teacher model is introduced to supervise the student output along with true labels, with its updates based on the moving average of the student model. 

In summary, the complete objective is formulated as:
\begin{equation}
\begin{aligned}
    \mathcal{L} = \mathcal{L}_{cla} + &\alpha \cdot\mathcal{L}_{con} + \beta \cdot \mathcal{L}_{comm}, \\
 \hspace{-3mm}   \text{with} \ \mathcal{L}_{cla}=-y\log(\sigma(\hat{y}_s))&, \mathcal{L}_{con} = KL(\sigma(\hat{y}_s/T)||\sigma(\hat{y}_t/T))
\end{aligned}
\end{equation}
where $\alpha$ and $\beta$ control the relative importance of each loss, $T$ is the temperature, $\sigma$ denotes the softmax activation, and $\mathcal{L}_{cla}$ and $\mathcal{L}_{con}$ denote the classification loss and consistency loss with the teacher model, respectively.

\begin{table}[t]
    \centering
    \setlength{\tabcolsep}{3.5pt}
    \caption{Generalzation results on DG benchmarks with ResNet-18.}
    \begin{tabular}{c|ccc|c}
    \toprule
    \multirow{2}{*}{Method} & \multicolumn{3}{c|}{Dataset} & \multirow{2}{*}{Avg.($\uparrow$)}\\
    \cmidrule(r){2-4}
     & PACS($\uparrow$) & Terra($\uparrow$) & VLCS($\uparrow$) & \\
    \midrule
    VREx \cite{krueger2021out} (ICML'2021)    & 80.97         & 38.60        & 76.62          & 65.40 \\
    MTL \cite{blanchard2021domain} (JMLR'2021)   & 80.60         & 40.55       & 75.38        & 65.51  \\
    SagNet \cite{nam2021reducing} (CVPR'2021)  & 81.55          & 38.75         & 76.24          & 65.51 \\
    ARM \cite{zhang2021adaptive} (NeurIPS'2021)     & 80.98         & 37.47        & 76.61        & 65.02 \\
    SAM \cite{foret2021sharpnessaware} (ICLR'2021)   & 82.35         & 41.76         & 76.45       & 66.85 \\
    FACT \cite{xu2021fourier} (CVPR'2021)     & 83.07          & 43.87         & 77.00        & 67.98\\
    SWAD \cite{cha2021swad} (NeurIPS'2021)    & 83.11        & 42.93         & 76.60        & 67.55 \\
    MIRO \cite{cha2022domain} (ECCV'2022)  & 79.28  & 42.63           & 76.38        & 66.10\\
    PCL \cite{yao2022pcl} (CVPR'2022)     & 82.63           & 43.21         & 76.32        & 67.39\\
    AdaNPC \cite{zhang2023adanpc} (ICML'2023) & 82.50          & 41.35         & 75.98        & 66.61\\
    DandelionNet \cite{hu2023dandelionnet} (ICCV'2023) & 82.34           & 41.98         & 74.08        & 66.13\\
    iDAG \cite{huang2023idag} (ICCV'2023) & 83.21         & 41.93         & 74.34       & 66.49\\
       SAGM \cite{wang2023sharpness}(CVPR'2023) & 81.34 & 40.66 & 75.83 & 65.94 \\
       GMDG \cite{tan2024rethinking} (CVPR'2024)& 81.71  & 43.34 & 75.81 & 66.95\\
    \midrule
    {DDG (ours)} & \makecell{\bfseries84.47\\$\pm$0.18}           & \makecell{\bfseries46.63\\$\pm$0.25}         & \makecell{\bfseries78.36\\$\pm$0.24} & \bfseries69.82 \\ 
    \bottomrule
    \end{tabular}
    \vspace{-2mm}
    \label{benchmarks}
    \vspace{-3mm}
\end{table}


\section{Experiments}
\begin{figure*}[t]
    \centering
    \includegraphics[width=\textwidth]{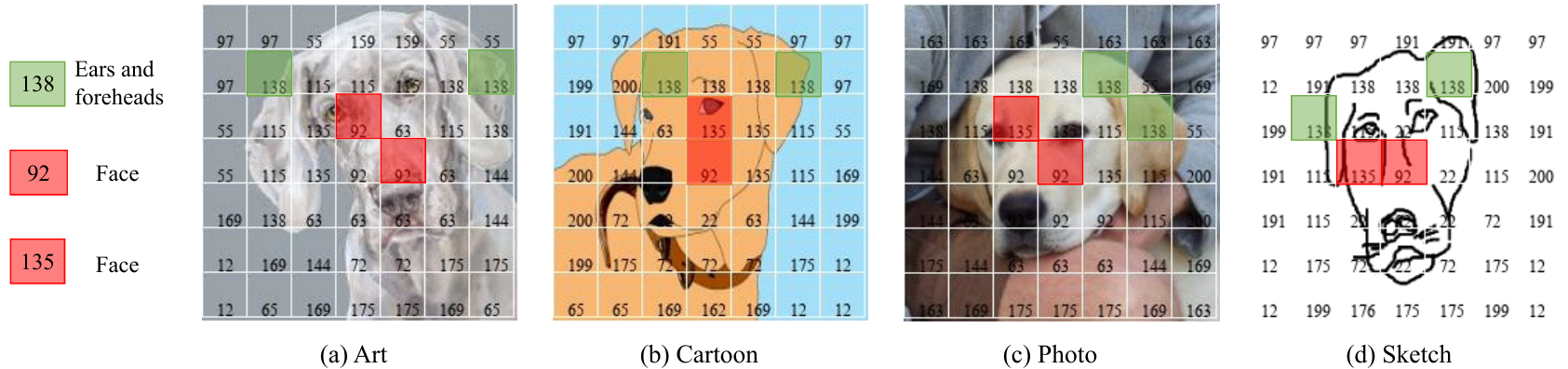}
    \vspace{-5mm}
    \caption{Visualizations illustrating the semantics of learned codewords in our DDG. The labels within the patches indicate the index of the codeword within the codebook. In our proposed DDG, patches in the feature maps are substituted with the corresponding codewords according to their respective indices.}
    \label{visua_code}
\end{figure*}

\subsection{Experiment Setup}

Following the common practice in DG~\cite{gulrajani2020search_2, zhou2024mixstyle, li2023sparse, long2024dgmamba}, we conducted experiments on widely used benchmarks: PACS~\cite{li2017deeper}, TerraIncognita (Terra)~\cite{beery2018recognition}, and VLCS~\cite{fang2013unbiased}, which encompass images sourced from diverse media~\cite{xie2023adapt}.
The ImageNet~\cite{deng2009imagenet} pre-trained ResNet-18~\cite{he2016deep} serves as the backbone for all experiments. Our approach involves training for 5k iterations using SGD, with a batch size of 32 and weight decay of 5e-4. The learning rates are initially set to 0.004 for PACS and Terra, and 0.001 for VLCS. with a 0.1 decay at 80\% of the total iterations. Our DDG is built on FACT~\cite{xu2021fourier}. The parameter $\alpha$ is set to 200 for Terra and VLCS, and 2 for PACS. While the weight~$\beta$ remains fixed at 0.1, and $T$ is set to 10 across all experiments. We report performance following the training-domain validation protocol~\cite{gulrajani2020search_2, long2024rethinking}. In this work, the total number of codewords is fixed as 256.

\subsection{Experiment Results}

\begin{figure}[t]
\vspace{-2mm}
  \centering
  \subfigure[Before discretization]{\includegraphics[width=0.21\textwidth]{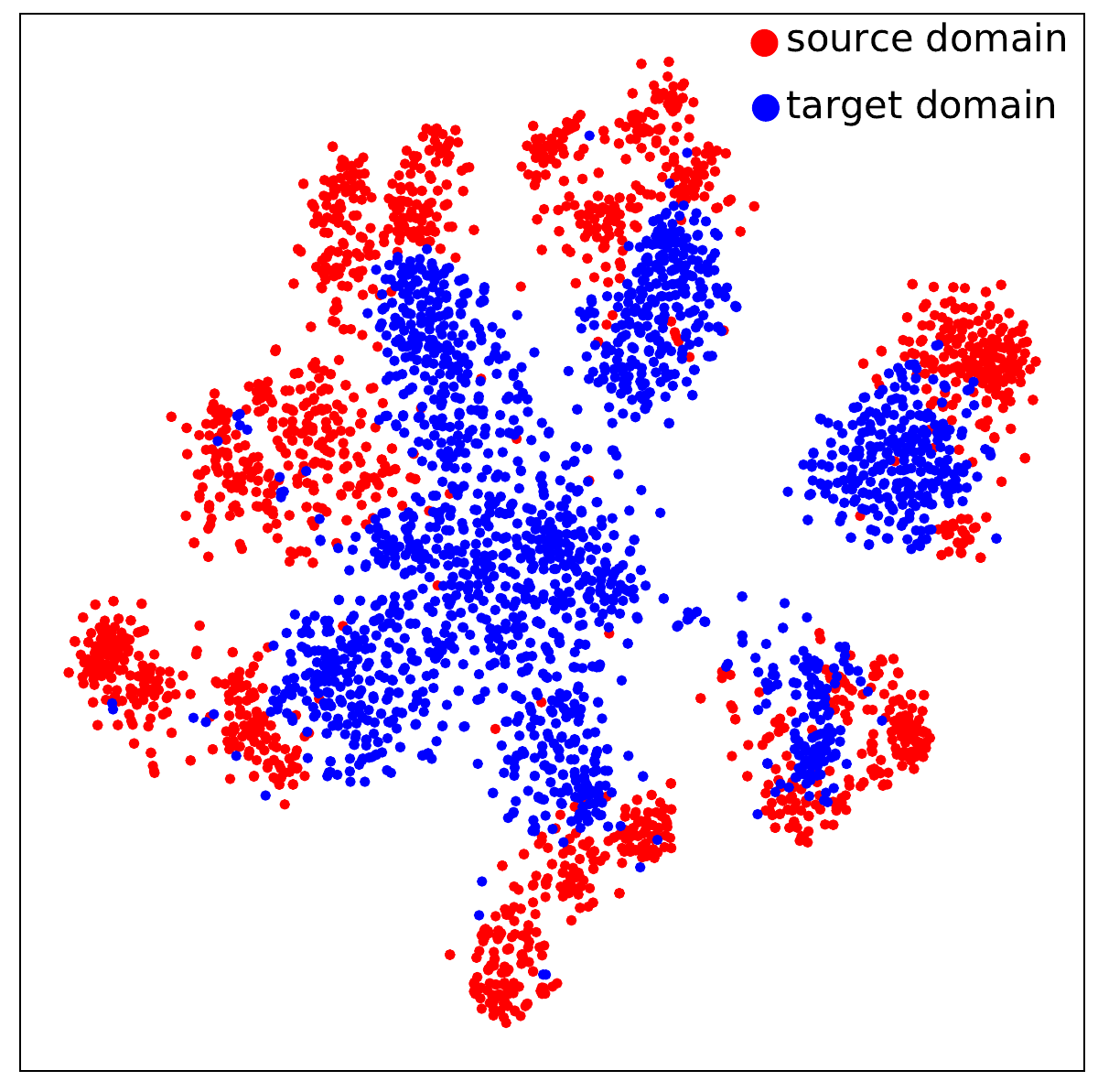}}
  \subfigure[After discretization]{\includegraphics[width=0.21\textwidth]{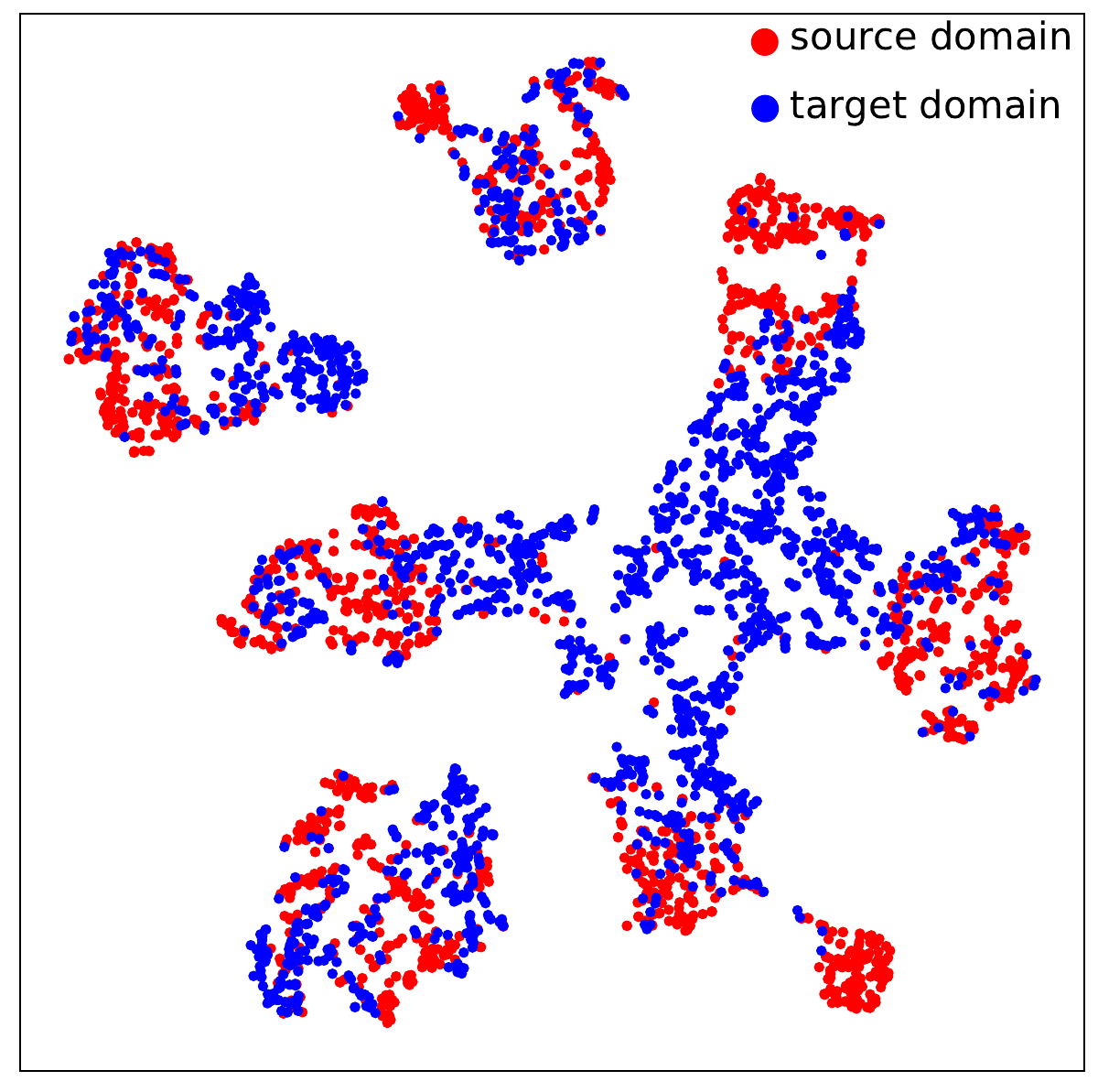}}
  \vspace{-2.5mm}
\caption{Visualization with t-SNE embeddings drawing features from the source and target domains before and after employing our DDG.}
  \label{ST}
  \vspace{-4mm}
\end{figure}

\noindent\textbf{Comparisons.}
The comparison with SOTA methods is reported in TABLE~\ref{benchmarks}. As shown, our DDG maintains enhancements of 9.6\%, 6.4\%, and 2.3\% over SOTA methods on PACS, Terra, and VLCS, respectively, consistently attaining the highest average recognition accuracy across diverse benchmarks. This underscores its supremacy in capturing essential semantic information. Despite the heightened difficulty posed by scene-centric images in Terra and VLCS for DG, DDG achieves the top performance. These outcomes underscore that our DDG, utilizing vector quantization, enhances generalizability by prioritizing semantic information over pixel-level details.

\noindent\textbf{Ablation Study.}
To assess the contributions of each component in the discretization, we perform ablation studies on PACS. The results in TABLE~\ref{ablation} underscore the efficacy and indispensability of both the VQ loss and the commitment loss for learning the powerful codewords. Moreover, optimizing the VQ loss using EMA achieves a 2.0\% improvement over the gradient descent strategy. This outcome supports the claim that incorporating EMA fosters a more resilient codebook, thereby improving the model's generalizability.

\begin{table}[t]
    \centering
    \setlength{\tabcolsep}{2pt}
    \caption{Ablation study of components on PACS.}
    \vspace{-1mm}{
    \begin{tabular}{c|ccc|cccc|c}
    \toprule
    \multirow{2}{*}{ID} & \multirow{2}{*}{$\mathcal{L}_{comm}$} & \multirow{2}{*}{\makecell{$\mathcal{L}_{vq}$ \\(via SGD)}} & \multirow{2}{*}{\makecell{$\mathcal{L}_{vq}$ \\(via EMA)}} & \multicolumn{4}{c|}{Target domain} & \multirow{2}{*}{Avg.($\uparrow$)}\\
    \cmidrule(r){5-8}
    &&&& Art & Cartoon & Photo & Sketch & \\
    \midrule
    \Rmnum{1}                & -          & -           & -  &79.93&75.43&92.28& 79.16    & 81.71 \\
     \Rmnum{2}      & \checkmark        & -         & -   &80.76&76.92&92.63&  79.38   & 82.42\\
     \Rmnum{3}       & -        & \checkmark         & -   &81.98&77.60&93.35& 76.23    & 82.40\\
     \Rmnum{4}      & -        & -         & \checkmark  &83.06&78.50&92.63&  79.56    & 83.44\\
     \Rmnum{5}       & \checkmark        & \checkmark         & - &82.61 &75.81 &92.16 & 80.73      & 82.83\\
    \Rmnum{6}          & \checkmark          & -           & \checkmark  &82.81 &78.41 & 94.07& 82.62       & \bfseries84.48 \\
    \bottomrule
    \end{tabular}} 
    \vspace{-5mm}
    \label{ablation}
\end{table}

\noindent\textbf{Visualization for Distribution Discrepancy.}
The finite codewords in the codebook constrain the feature space, facilitating domain alignment compared to continuous representation learning. Fig.~\ref{ST} displays t-SNE embeddings\cite{van2008visualizing} of features from source and target domains, showing that the discrete codebook results in a closer alignment of the distributions. Our DDG lacks an explicit alignment strategy, yet a noticeable reduction in distribution gaps is observed, suggesting that the DDG captures essential semantic features rather than pixel details, reducing the difficulty of mitigating distribution gaps.

\noindent\textbf{Visualization for Codeword Semantics.}
To understand the semantics of codewords and their role in domain alignment, we track codeword selections for each patch~(an image is processed into 7~$\times$~7 patches), and visualize the discrete features across domains from PACS. Fig.~\ref{visua_code} shows that the learned codewords for patches with similar semantics are consistent across domains, despite differences in styles and shapes. For instance, codeword 138 represents \textit{dogs' ears and foreheads}, while codewords 92 and 135 capture \textit{dog's faces}. This codeword consistency between similar semantics across domains simplifies distribution alignment, demonstrating DDG's efficacy in learning at the semantic level rather than the pixel level and reducing distribution gaps across domains.

\begin{table}[t]
    \centering 
    \vspace{-5mm}
    \setlength{\tabcolsep}{1pt}
    \caption{Generalization stability on DG benchmarks. }
    {
    \begin{tabular}{c|ccc|c}
    \toprule
    \multirow{2}{*}{Method} & \multicolumn{3}{c|}{dataset} & \multirow{2}{*}{Avg.($\downarrow$)}\\
    \cmidrule(r){2-4}
     & PACS($\downarrow$)& Terra($\downarrow$)& VLCS($\downarrow$)\\
    \midrule
    FACT \cite{xu2021fourier}  (CVPR'2021)        & 8.32           & 10.51        & 14.33       & 11.05 \\
     SWAD \cite{cha2021swad}  (NeurIPS'2021)        & 9.65           & 11.51        & 15.44      & 12.20 \\
    PCL \cite{yao2022pcl}  (CVPR'2022)         & 9.92           & 9.28          &14.87       & 11.36\\
    MIRO \cite{cha2022domain}  (CVPR'2022)         & 13.38           &11.70          &15.23       & 13.44\\
     DandelionNet \cite{hu2023dandelionnet}  (ICCV'2023)  & 9.40           & 11.31       & \bfseries14.11       & 11.61\\
     iDAG \cite{huang2023idag}  (ICCV'2023)  & 9.66           & 12.00        & 14.52       & 12.06\\
       SAGM \cite{wang2023sharpness} (CVPR'2023) & 10.14 & 12.92 & 14.78 & 12.61 \\
       GMDG \cite{tan2024rethinking} (CVPR'2024)& 12.60  & 8.85 & 14.46& 11.97\\
    \midrule
    {DDG (ours)}       & \bfseries6.71           & \bfseries8.62          & 14.68         & \bfseries10.08 \\
    \bottomrule
    \end{tabular}}
    \vspace{-5mm}
    \label{GS}
\end{table}

\noindent\textbf{Generalization Stability.}
The robustness of generalizability across diverse scenarios is an essential metric to discern whether the model overfits in easy-to-transfer domains and struggles in hard-to-transfer domains~\cite{long2023diverse}. Following the protocol in~\cite{long2023diverse}, we utilize the generalization stability~($GS$) metric to assess the robustness of models' generalizability and report it in Table~\ref{GS}. 
The lowest $GS$ values in PACS and TerraIncognita underscore the superiority of our DDG in enhancing the generalization stability. Although the $GS$ value of DDG in VLCS does not beat the SOTA method, it is worth noting that the difference is insignificant.

For the definition of the metric $GS$ and additional experimental results, please refer to the supplementary material.

\section{Conclusion}

This paper introduces a pioneering paradigm for DG by approaching it through the lens of discrete representation codebook learning. We theoretically illustrate the excellence of discrete codewords in reducing distribution gaps compared to prevailing continuous representation learning approaches. Motivated by this insight, our proposed framework, named Discrete Domain Generalization (DDG), quantizes continuous features into discrete codewords, aiming to capture essential semantic features at a semantic level rather than the conventional pixel level. This approach reduces the number of latent variables and aids in domain alignment. Comprehensive experiments on commonly used benchmarks demonstrate the effectiveness and superiority of our DDG, highlighting a new direction for enhancing the model's generalizability through discrete representation learning.

\bibliographystyle{IEEEbib}
\bibliography{DG}

\begin{thebibliography}{10}

\bibitem{ganin2016domainadversarial}
Yaroslav Ganin, Evgeniya Ustinova, Hana Ajakan, Pascal Germain, Hugo Larochelle, Fran{\c c}ois Laviolette, Mario March, and Victor Lempitsky,
\newblock ``Domain-{{adversarial training}} of {{neural networks}},''
\newblock {\em JMLR}, vol. 17, no. 59, pp. 1--35, 2016.

\bibitem{long2024rethinking}
Shaocong Long, Qianyu Zhou, Chenhao Ying, Lizhuang Ma, and Yuan Luo,
\newblock ``Rethinking domain generalization: Discriminability and generalizability,''
\newblock {\em TCSVT}, pp. 11783--11797, 2024.

\bibitem{zhao2020domain}
Shanshan Zhao, Mingming Gong, Tongliang Liu, Huan Fu, and Dacheng Tao,
\newblock ``Domain {{generalization}} via {{entropy regularization}},''
\newblock {\em NeurIPS}, vol. 33, pp. 16096--16107, 2020.

\bibitem{zhou2024mixstyle}
Kaiyang Zhou, Yongxin Yang, Yu~Qiao, and Tao Xiang,
\newblock ``Mixstyle neural networks for domain generalization and adaptation,''
\newblock {\em IJCV}, vol. 132, no. 3, pp. 822--836, 2024.

\bibitem{zhao2024style}
Yuyang Zhao, Zhun Zhong, Na~Zhao, Nicu Sebe, and Gim~Hee Lee,
\newblock ``Style-hallucinated dual consistency learning: A unified framework for visual domain generalization,''
\newblock {\em IJCV}, vol. 132, no. 3, pp. 837--853, 2024.

\bibitem{xu2021fourier}
Qinwei Xu, Ruipeng Zhang, Ya~Zhang, Yanfeng Wang, and Qi~Tian,
\newblock ``A fourier-based framework for domain generalization,''
\newblock in {\em CVPR}, 2021, pp. 14383--14392.

\bibitem{yang2024pointdgmamba}
Hao Yang, Qianyu Zhou, Haijia Sun, Xiangtai Li, Fengqi Liu, Xuequan Lu, Lizhuang Ma, and Shuicheng Yan,
\newblock ``Pointdgmamba: Domain generalization of point cloud classification via generalized state space model,''
\newblock in {\em AAAI}, 2025.

\bibitem{zhou2023instance}
Qianyu Zhou, Ke-Yue Zhang, Taiping Yao, Xuequan Lu, Ran Yi, Shouhong Ding, and Lizhuang Ma,
\newblock ``Instance-aware domain generalization for face anti-spoofing,''
\newblock in {\em CVPR}, 2023, pp. 20453--20463.

\bibitem{wang2022domain}
Zhuo Wang, Zezheng Wang, Zitong Yu, Weihong Deng, Jiahong Li, Tingting Gao, and Zhongyuan Wang,
\newblock ``Domain generalization via shuffled style assembly for face anti-spoofing,''
\newblock in {\em CVPR}, 2022, pp. 4123--4133.

\bibitem{zhang2022principled}
Hanlin Zhang, Yi-Fan Zhang, Weiyang Liu, Adrian Weller, Bernhard Sch{\"o}lkopf, and Eric~P Xing,
\newblock ``Towards principled disentanglement for domain generalization,''
\newblock in {\em CVPR}, 2022, pp. 8024--8034.

\bibitem{yao2022pcl}
Xufeng Yao, Yang Bai, Xinyun Zhang, Yuechen Zhang, Qi~Sun, Ran Chen, Ruiyu Li, and Bei Yu,
\newblock ``Pcl: Proxy-based contrastive learning for domain generalization,''
\newblock in {\em CVPR}, 2022, pp. 7097--7107.

\bibitem{kim2021selfreg}
Daehee Kim, Youngjun Yoo, Seunghyun Park, Jinkyu Kim, and Jaekoo Lee,
\newblock ``Selfreg: Self-supervised contrastive regularization for domain generalization,''
\newblock in {\em ICCV}, 2021, pp. 9619--9628.

\bibitem{cha2021swad}
Junbum Cha, Sanghyuk Chun, Kyungjae Lee, Han-Cheol Cho, Seunghyun Park, Yunsung Lee, and Sungrae Park,
\newblock ``Swad: Domain generalization by seeking flat minima,''
\newblock {\em NeurIPS}, vol. 34, pp. 22405--22418, 2021.

\bibitem{wang2023sharpness}
Pengfei Wang, Zhaoxiang Zhang, Zhen Lei, and Lei Zhang,
\newblock ``Sharpness-aware gradient matching for domain generalization,''
\newblock in {\em CVPR}, 2023, pp. 3769--3778.

\bibitem{li2023sparse}
Bo~Li, Yifei Shen, Jingkang Yang, Yezhen Wang, Jiawei Ren, Tong Che, Jun Zhang, and Ziwei Liu,
\newblock ``Sparse mixture-of-experts are domain generalizable learners,''
\newblock in {\em ICLR}, 2023.

\bibitem{zhou2022adaptive}
Qianyu Zhou, Ke-Yue Zhang, Taiping Yao, Ran Yi, Shouhong Ding, and Lizhuang Ma,
\newblock ``Adaptive mixture of experts learning for generalizable face anti-spoofing,''
\newblock in {\em ACMMM}, 2022, pp. 6009--6018.

\bibitem{jiang2024dgpic}
Jincen Jiang, Qianyu Zhou, Yuhang Li, Xuequan Lu, Meili Wang, Lizhuang Ma, Jian Chang, and Jian~Jun Zhang,
\newblock ``Dg-pic: Domain generalized point-in-context learning for point cloud understanding,''
\newblock in {\em ECCV}. Springer, 2024, pp. 455--474.

\bibitem{zhou2024test}
Qianyu Zhou, Ke-Yue Zhang, Taiping Yao, Xuequan Lu, Shouhong Ding, and Lizhuang Ma,
\newblock ``Test-time domain generalization for face anti-spoofing,''
\newblock in {\em CVPR}, 2024, pp. 175--187.

\bibitem{van2017neural}
Aaron Van Den~Oord, Oriol Vinyals, et~al.,
\newblock ``Neural discrete representation learning,''
\newblock {\em NeurIPS}, vol. 30, pp. 6306--6315, 2017.

\bibitem{esser2021taming}
Patrick Esser, Robin Rombach, and Bjorn Ommer,
\newblock ``Taming transformers for high-resolution image synthesis,''
\newblock in {\em CVPR}, 2021, pp. 12873--12883.

\bibitem{radford2021learning}
Alec Radford, Jong~Wook Kim, Chris Hallacy, Aditya Ramesh, Gabriel Goh, Sandhini Agarwal, Girish Sastry, Amanda Askell, Pamela Mishkin, Jack Clark, et~al.,
\newblock ``Learning transferable visual models from natural language supervision,''
\newblock in {\em ICML}, 2021, pp. 8748--8763.

\bibitem{sriperumbudur2012empirical}
Bharath~K. Sriperumbudur, Kenji Fukumizu, Arthur Gretton, Bernhard Sch{\"o}lkopf, and Gert R.~G. Lanckriet,
\newblock ``{On the empirical estimation of integral probability metrics},''
\newblock {\em Electronic Journal of Statistics}, vol. 6, pp. 1550 -- 1599, 2012.

\bibitem{ben2010theory}
Shai Ben-David, John Blitzer, Koby Crammer, Alex Kulesza, Fernando Pereira, and Jennifer~Wortman Vaughan,
\newblock ``A theory of learning from different domains,''
\newblock {\em Machine Learning}, vol. 79, no. 1, pp. 151--175, 2010.

\bibitem{krueger2021out}
David Krueger, Ethan Caballero, Joern-Henrik Jacobsen, Amy Zhang, Jonathan Binas, Dinghuai Zhang, Remi Le~Priol, and Aaron Courville,
\newblock ``Out-of-distribution generalization via risk extrapolation (rex),''
\newblock in {\em ICML}, 2021, pp. 5815--5826.

\bibitem{blanchard2021domain}
Gilles Blanchard, Aniket~Anand Deshmukh, {\"U}run Dogan, Gyemin Lee, and Clayton Scott,
\newblock ``Domain generalization by marginal transfer learning,''
\newblock {\em JMLR}, vol. 22, no. 1, pp. 46--100, 2021.

\bibitem{nam2021reducing}
Hyeonseob Nam, HyunJae Lee, Jongchan Park, Wonjun Yoon, and Donggeun Yoo,
\newblock ``Reducing domain gap by reducing style bias,''
\newblock in {\em CVPR}, 2021, pp. 8690--8699.

\bibitem{zhang2021adaptive}
Marvin Zhang, Henrik Marklund, Nikita Dhawan, Abhishek Gupta, Sergey Levine, and Chelsea Finn,
\newblock ``Adaptive risk minimization: Learning to adapt to domain shift,''
\newblock {\em NeurIPS}, vol. 34, pp. 23664--23678, 2021.

\bibitem{foret2021sharpnessaware}
Pierre Foret, Ariel Kleiner, Hossein Mobahi, and Behnam Neyshabur,
\newblock ``Sharpness-aware minimization for efficiently improving generalization,''
\newblock in {\em ICLR}, 2021.

\bibitem{cha2022domain}
Junbum Cha, Kyungjae Lee, Sungrae Park, and Sanghyuk Chun,
\newblock ``Domain generalization by mutual-information regularization with pre-trained models,''
\newblock in {\em ECCV}, 2022, pp. 440--457.

\bibitem{zhang2023adanpc}
Yifan Zhang, Xue Wang, Kexin Jin, Kun Yuan, Zhang Zhang, Liang Wang, Rong Jin, and Tieniu Tan,
\newblock ``Adanpc: Exploring non-parametric classifier for test-time adaptation,''
\newblock in {\em ICML}, 2023, pp. 41647--41676.

\bibitem{hu2023dandelionnet}
Lanqing Hu, Meina Kan, Shiguang Shan, and Xilin Chen,
\newblock ``Dandelionnet: Domain composition with instance adaptive classification for domain generalization,''
\newblock in {\em ICCV}, 2023, pp. 19050--19059.

\bibitem{huang2023idag}
Zenan Huang, Haobo Wang, Junbo Zhao, and Nenggan Zheng,
\newblock ``idag: Invariant dag searching for domain generalization,''
\newblock in {\em ICCV}, 2023, pp. 19169--19179.

\bibitem{tan2024rethinking}
Zhaorui Tan, Xi~Yang, and Kaizhu Huang,
\newblock ``Rethinking multi-domain generalization with a general learning objective,''
\newblock in {\em CVPR}, 2024, pp. 23512--23522.

\bibitem{gulrajani2020search_2}
Ishaan Gulrajani and David Lopez-Paz,
\newblock ``In search of lost domain generalization,''
\newblock in {\em ICLR}, 2020.

\bibitem{long2024dgmamba}
Shaocong Long, Qianyu Zhou, Xiangtai Li, Xuequan Lu, Chenhao Ying, Yuan Luo, Lizhuang Ma, and Shuicheng Yan,
\newblock ``Dgmamba: Domain generalization via generalized state space model,''
\newblock in {\em ACMMM}, 2024, pp. 3607--3616.

\bibitem{li2017deeper}
Da~Li, Yongxin Yang, Yi-Zhe Song, and Timothy~M. Hospedales,
\newblock ``Deeper, {{broader}} and {{artier domain generalization}},''
\newblock in {\em ICCV}, 2017, pp. 5543--5551.

\bibitem{beery2018recognition}
Sara Beery, Grant Van~Horn, and Pietro Perona,
\newblock ``Recognition in terra incognita,''
\newblock in {\em ECCV}, 2018, pp. 456--473.

\bibitem{fang2013unbiased}
Chen Fang, Ye~Xu, and Daniel~N Rockmore,
\newblock ``Unbiased metric learning: On the utilization of multiple datasets and web images for softening bias,''
\newblock in {\em ICCV}, 2013, pp. 1657--1664.

\bibitem{xie2023adapt}
Han Xie, Zhifeng Shen, Shicai Yang, Weijie Chen, and Luojun Lin,
\newblock ``Adapt then generalize: A simple two-stage framework for semi-supervised domain generalization,''
\newblock in {\em ICME}, 2023, pp. 540--545.

\bibitem{deng2009imagenet}
Jia Deng, Wei Dong, Richard Socher, Li-Jia Li, Kai Li, and Li~Fei-Fei,
\newblock ``Imagenet: A large-scale hierarchical image database,''
\newblock in {\em CVPR}, 2009, pp. 248--255.

\bibitem{he2016deep}
Kaiming He, Xiangyu Zhang, Shaoqing Ren, and Jian Sun,
\newblock ``Deep {{residual learning}} for {{image recognition}},''
\newblock in {\em CVPR}, 2016, pp. 770--778.

\bibitem{van2008visualizing}
Laurens Van~der Maaten and Geoffrey Hinton,
\newblock ``Visualizing data using t-sne,''
\newblock {\em JMLR}, vol. 9, no. 11, pp. 2579--2605, 2008.

\bibitem{long2023diverse}
Shaocong Long, Qianyu Zhou, Chenhao Ying, Lizhuang Ma, and Yuan Luo,
\newblock ``Diverse target and contribution scheduling for domain generalization,''
\newblock {\em arXiv preprint arXiv:2309.16460}, 2023.

\bibitem{wang2022semantic}
Mengzhu Wang, Jianlong Yuan, Qi~Qian, Zhibin Wang, and Hao Li,
\newblock ``Semantic data augmentation based distance metric learning for domain generalization,''
\newblock in {\em ACMMM}, 2022, pp. 3214--3223.

\bibitem{zhou2020learning}
Kaiyang Zhou, Yongxin Yang, Timothy Hospedales, and Tao Xiang,
\newblock ``Learning to generate novel domains for domain generalization,''
\newblock in {\em ECCV}. Springer, 2020, pp. 561--578.

\bibitem{li2018domain}
Ya~Li, Mingming Gong, Xinmei Tian, Tongliang Liu, and Dacheng Tao,
\newblock ``Domain {{generalization}} via {{conditional invariant representations}},''
\newblock {\em Proceedings of the AAAI Conference on Artificial Intelligence}, vol. 32, no. 1, 2018.

\bibitem{zhang2023free}
YiFan Zhang, xue wang, Jian Liang, Zhang Zhang, Liang Wang, Rong Jin, and Tieniu Tan,
\newblock ``Free lunch for domain adversarial training: Environment label smoothing,''
\newblock in {\em ICLR}, 2023.

\bibitem{liu2021domain}
Chang Liu, Lichen Wang, Kai Li, and Yun Fu,
\newblock ``Domain generalization via feature variation decorrelation,''
\newblock in {\em ACMMM}, 2021, pp. 1683--1691.

\bibitem{wang2022variational}
Yufei Wang, Haoliang Li, Hao Cheng, Bihan Wen, Lap-Pui Chau, and Alex Kot,
\newblock ``Variational disentanglement for domain generalization,''
\newblock {\em TMLR}, 2022.

\bibitem{huang2023sentence}
Zeyi Huang, Andy Zhou, Zijian Ling, Mu~Cai, Haohan Wang, and Yong~Jae Lee,
\newblock ``A sentence speaks a thousand images: Domain generalization through distilling clip with language guidance,''
\newblock in {\em ICCV}, 2023, pp. 11685--11695.

\bibitem{mahajan2021domain}
Divyat Mahajan, Shruti Tople, and Amit Sharma,
\newblock ``Domain generalization using causal matching,''
\newblock in {\em ICML}. PMLR, 2021, pp. 7313--7324.

\bibitem{chu2022dna}
Xu~Chu, Yujie Jin, Wenwu Zhu, Yasha Wang, Xin Wang, Shanghang Zhang, and Hong Mei,
\newblock ``{{DNA}}: {{Domain generalization}} with diversified neural averaging,''
\newblock in {\em Proceedings of the {{International Conference}} on {{Machine Learning}}}. 2022, pp. 4010--4034, {PMLR}.

\bibitem{sultana2022self}
Maryam Sultana, Muzammal Naseer, Muhammad~Haris Khan, Salman Khan, and Fahad~Shahbaz Khan,
\newblock ``Self-distilled vision transformer for domain generalization,''
\newblock in {\em ACCV}, 2022, pp. 3068--3085.

\bibitem{guo2024seta}
Jintao Guo, Lei Qi, Yinghuan Shi, and Yang Gao,
\newblock ``Seta: Semantic-aware edge-guided token augmentation for domain generalization,'' 2024.

\end{thebibliography}

\clearpage
\appendix

In this supplementary material, we present additional content to further demonstrate the advantages of our proposed DDG, including an overview of related work and supplementary comparison experiments.

\section{Related Work}
Numerous research endeavors in DG have been devoted to augmenting model generalizability in novel scenarios.

Mainstream approaches in DG have primarily centered on maintaining domain-invariant representation with the aim of achieving powerful expressive representation. Data augmentation~\cite{zhou2024mixstyle, wang2022semantic, zhou2020learning, zhao2024style} augment the source domain with more generated data exhibiting diverse styles, aiming to expose the model to a broader range of scenarios. Distribution alignment~\cite{ganin2016domainadversarial, li2018domain, zhao2020domain, zhang2023free} employs domain adversarial training to remove domain-specific information. As an effective way to capture the genuine semantic features, disentangle techniques~\cite{liu2021domain, wang2022variational, wang2022domain, zhang2022principled} seek to disentangle features into semantic and non-semantic information. Contrastive learning~\cite{yao2022pcl, kim2021selfreg, huang2023sentence, mahajan2021domain} introduces contrastive loss to regularize the acquired features to be close to those with the same label. Inspired by the generalization performance of flatness-aware strategy, stochastic weight averaging~\cite{cha2021swad, wang2023sharpness, chu2022dna} attempts to find a flatter minima in loss landscapes. 
To alleviate the challenges associated with learning expressive representations through a single expert and complement the domain-shared information, methods employing a mixture-of-experts paradigm~\cite{li2023sparse, zhou2022adaptive} have been explored. These methods aim to mine sufficient and fine-grained information that may be absent in a single expert, releasing the constraints that a single expert may face when dealing with the substantial variability of data. 

Nevertheless, existing approaches to address distribution shifts in DG tend to acquire robust representation with continuous features and train at the pixel level, they grapple with challenges in the face of the expansive scope of continuous features. In contrast, our approach introduces discrete representation learning to obtain potent representations at the semantic level, with the goal of addressing the predicament posed by the vast space of continuous features while preserving crucial semantic features.



\section{Experiments}

\subsection{Implementation Details}
Following the common practice in DG research~\cite{gulrajani2020search_2, zhou2024mixstyle, li2023sparse}, we conducted experiments on widely used benchmarks, namely, PACS, TerraIncognita, and VLCS. These datasets encompass images sourced from diverse media, including hand-drawn illustrations, software-composited images, object-centered photographs, and scene-centered shots, thereby exhibiting substantial distribution shifts. Specifically, PACS includes 9991 images categorized into 7 classes, each exhibiting 4 diverse styles. TerraIncognita contains 24330 photographs of 10 kinds of wide animals captured at 4 distinct locations. VLCS is comprised of 4 sub-datasets, collectively consisting of 10729 images in 5 classes. 

The metric $GS$ is defined as $GS = \sqrt{\sum_{i = 1}^M(GP(i) - \overline{GP})^2},$ where $GP(i)$ denotes the generalization performance on domain $i$, and $\overline{GP} := \frac{1}{M}\sum_{i =1}^M GP(i)$ represents the average generalization performance across all domains.

\subsection{Additional Experiments}
\begin{figure}[t]
  \centering
  \subfigure[Before discretization]{\includegraphics[width=0.21\textwidth]{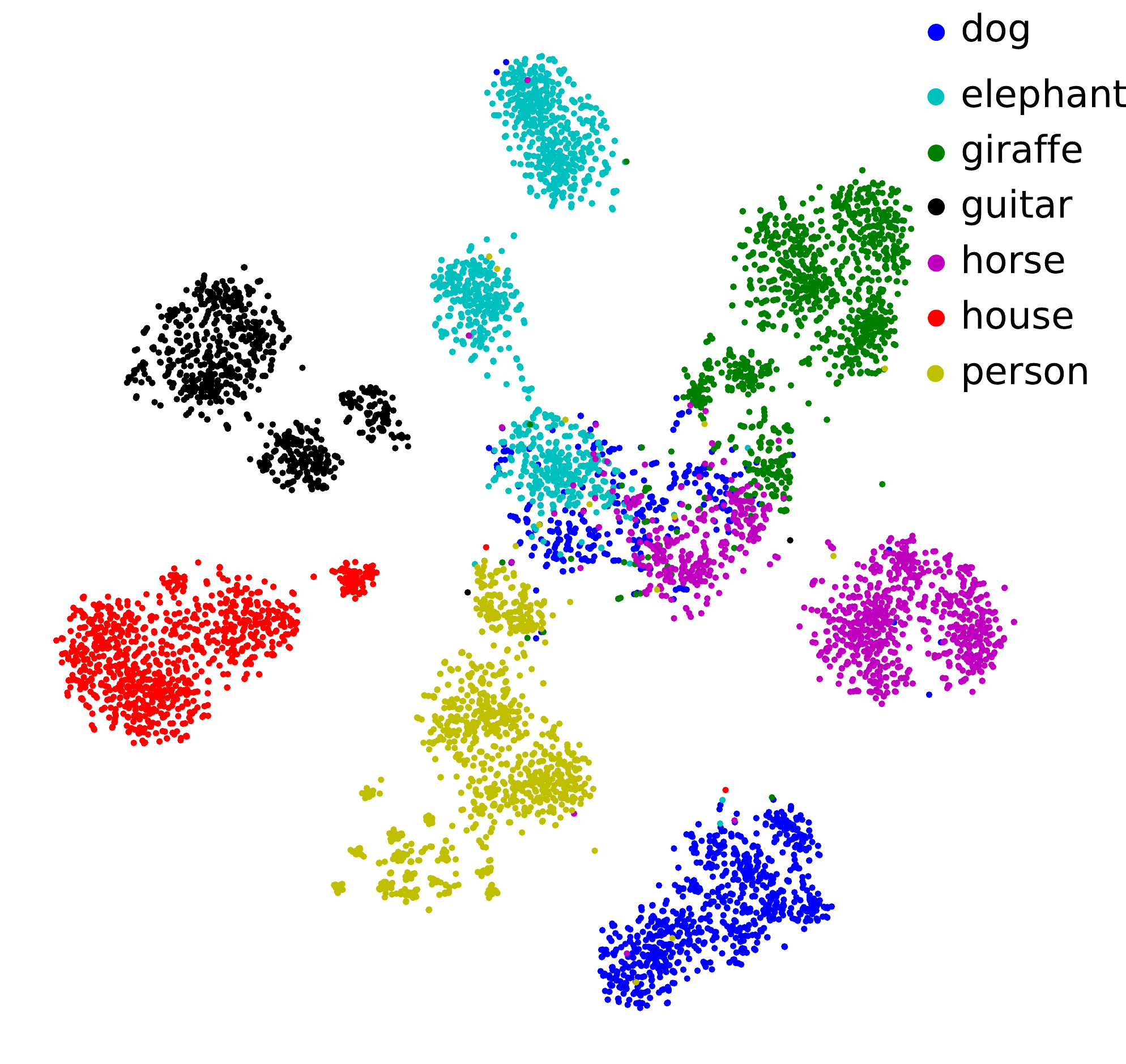}}
  \subfigure[After discretization]{\includegraphics[width=0.21\textwidth]{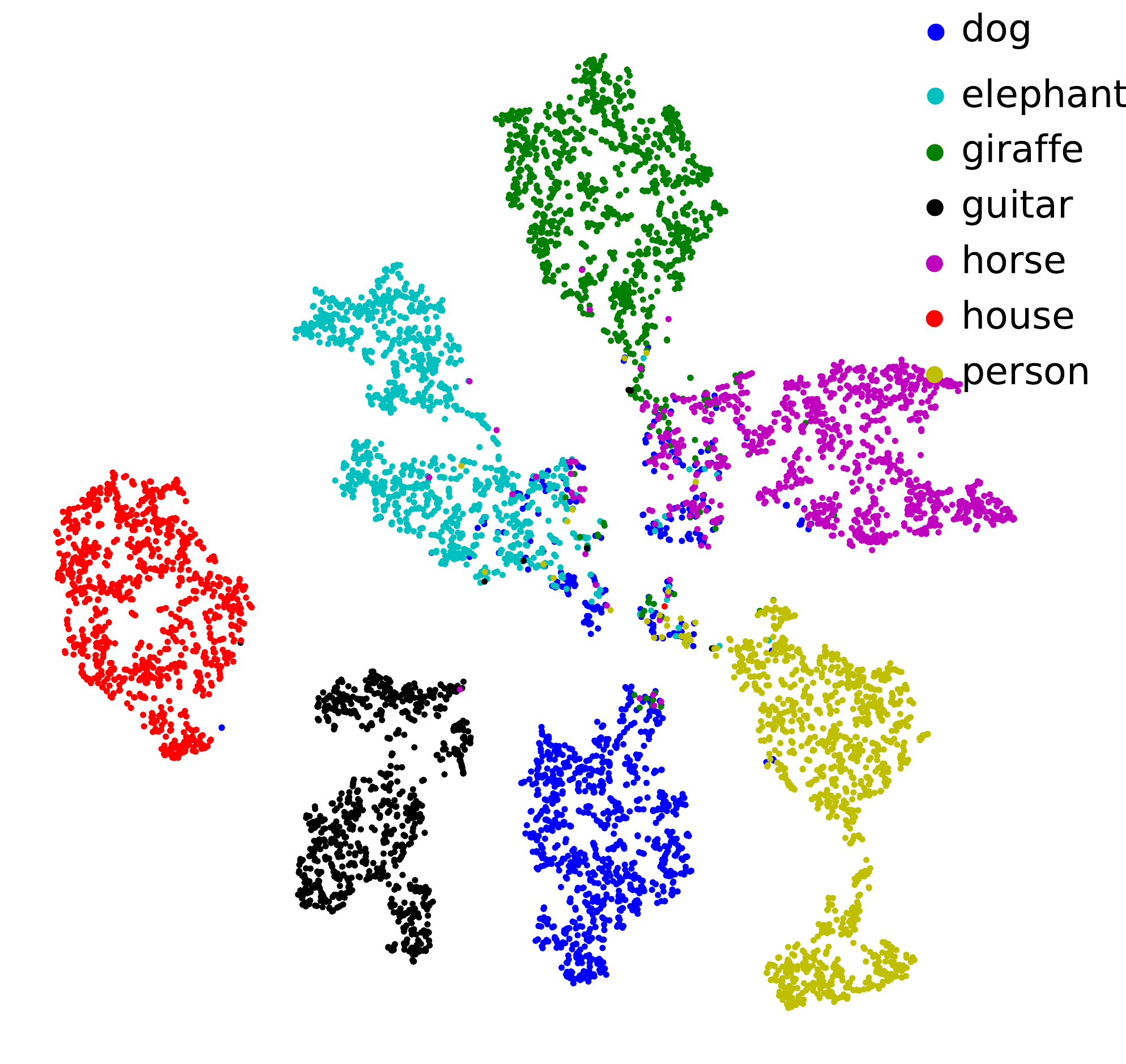}}
\vspace{-2mm}
\caption{Visualization with t-SNE embeddings depicting features from different classes before and after the application of the proposed DDG.}
  \label{visualiza}
  \vspace{2mm}
\end{figure}
\begin{table}[t]
\vspace{-3mm}
    \centering
    \caption{More comparisons on PACS. $\dag$ denotes reproduced result.}
    \vspace{-2mm}
    \setlength{\tabcolsep}{2pt}
    \begin{tabular}{c|c|c|c|c|c}
    \toprule
       Method & Venue & Avg.($\uparrow$) & Method & Venue & Avg.($\uparrow$)  \\
        \hline
        \multicolumn{3}{c|}{ResNet50} & \multicolumn{3}{c}{Deit-S} \\
        \midrule
        GMDG\cite{tan2024rethinking} & CVPR'2024 & 85.60 & SDViT\cite{sultana2022self}& ACCV'2022&86.3  \\
        \midrule
       SETA\cite{guo2024seta} & TIP'2024 & 87.18 & GMoE\cite{li2023sparse}$^{\dag}$& ICLR'2023&86.7  \\
        \midrule
        DDG (ours) &- &\textbf{87.29} & DDG (ours) &- & \textbf{87.1}  \\
        \bottomrule
    \end{tabular}
    \label{comparisons}
    \vspace{-3mm}
\end{table}

\noindent\textbf{Visualization for Class Separation.} To empirically substantiate the enhanced efficacy of the proposed DDG in encapsulating pivotal semantic features and acquiring robust representations, we utilize t-SNE embeddings~\cite{van2008visualizing} for the visualizations of acquired features from different classes both before and after the introduced vector quantization process. As illustrated in Figure.~\ref{visualiza}, the discernible augmentation in the separation of representations among distinct classes becomes evident subsequent to the discretization of features, such as the distinction between `elephant', `giraffe', and `horse'. Besides, the intra-class compactness, which could reflect domain gaps, becomes higher after the application of our DDG, as seen in the compactness in `person' and `giraffe'. The observations underscore the effectiveness of DDG in prioritizing the capture of semantic information over imperceptible pixel details and thereby acquiring powerful representations with high intra-class compactness and inter-class separation.
\begin{table}[t]
    \centering
    \setlength{\tabcolsep}{2.5pt}
    \caption{Generalization performance on PACS with object recognition accuracy~(\%) based on ResNet-18 pre-trained on ImageNet. High is better, and bold indicates the best performance.}
    \begin{tabular}{c|cccc|c}
    \toprule
    \multirow{2}{*}{Method} & \multicolumn{4}{c|}{Target domain} & \multirow{2}{*}{Avg.($\uparrow$)}\\
    \cmidrule(r){2-5}
     & Art & Cartoon & Photo & Sketch \\
    \midrule
    VREx \cite{krueger2021out} (ICML'2021)    & $80.84$         & $70.95$        & $93.64$         & 78.44          & $80.97$ \\
    MTL \cite{blanchard2021domain} (JMLR'2021)   & $79.99$         & $72.18$        & $95.28$         & $74.94$        & $80.60$  \\
    SagNet \cite{nam2021reducing} (CVPR'2021)  & 81.15           & $75.05$        & $94.61$         & 75.38          & $81.55$ \\
    ARM \cite{zhang2021adaptive} (NeurIPS'2021)     & $80.42$         & $75.96$        & $95.21$         & $72.33$        & $80.98$ \\
    SAM \cite{foret2021sharpnessaware} (ICLR'2021)   & $80.67$         & $75.53$        & 93.86         & $79.33$        & $82.35$ \\
    FACT \cite{xu2021fourier} (CVPR'2021)     & \bfseries84.08           & 75.30          & 94.31         & 78.57        & 83.07\\
    SWAD \cite{cha2021swad} (NeurIPS'2021)    & $83.28$         & $74.63$        & 96.56         & $77.96$        & $83.11$ \\
    MIRO \cite{cha2022domain} (ECCV'2022)  & 82.43  & 73.19          & 96.33           & 65.17         & 79.28\\
    PCL \cite{yao2022pcl} (CVPR'2022)     & 83.53           & 73.61          & $96.18$         & $77.20$        & $82.63$\\
    AdaNPC \cite{zhang2023adanpc} (ICML'2023) & 82.70           & 76.80          & 92.80         & 77.70        & 82.50\\
    DandelionNet \cite{hu2023dandelionnet} (ICCV'2023) & 83.16           & 74.36          & 95.28         & 76.56        & 82.34\\
    iDAG \cite{huang2023idag} (ICCV'2023) & 82.18           & 78.20          & 97.08         & 75.38       & 83.21\\
       SAGM \cite{wang2023sharpness}(CVPR'2023) & 81.76 & 74.68 & 95.51 & 73.41 & 81.34 \\
       GMDG \cite{tan2024rethinking} (CVPR'2024)& 83.77  & 75.64 & \bfseries97.38 & 67.91 & 81.71\\
    \midrule
    {DDG (ours)} & 82.75        & \bfseries78.71 & 93.83         & \bfseries82.62 & \bfseries84.47 \\ 
    \bottomrule
    \end{tabular}
    \label{PACS}
    \vspace{-3mm}
\end{table}

\noindent\textbf{Comparisons across Diverse Backbones.} 
In this section, we present additional comparisons with state-of-the-art models using various backbones, namely ResNet50 and ViT, as shown in Table \ref{comparisons}. The results demonstrate that our DDG consistently outperforms these models across different backbone architectures.

\noindent\textbf{Complexity Analysis.}
we provide the complexity comparison with ERM in Table~\ref{complexity}. As observed, the additional overhead of our DDG is negligible.
\begin{table}[h]
\vspace{-5mm}
    \centering
    \caption{Comparisons of computational efficiency.}
    \setlength{\tabcolsep}{3pt}
    \begin{tabular}{c|c|c|c}
    \toprule
       Backbone &Method &  Params & GFlops  \\
       \hline
       \multirow{3}{*}{ResNet50} & ERM & 23.5M & 4.1G \\
       \cmidrule(r){2-4}
       &                         DDG(ours) & 23.6M & 4.3G \\
        \bottomrule
    \end{tabular}
    \label{complexity}
    \vspace{-2mm}
\end{table}

\subsection{Full Results}

\begin{table}[t]
    \centering
    \setlength{\tabcolsep}{2.5pt}
    \caption{Generalizablity on TerraIncognita with object recognition accuracy~(\%) based on ResNet-18 pre-trained on ImageNet. High is better, and bold indicates the best performance.}
    \begin{tabular}{c|cccc|c}
    \toprule
    \multirow{2}{*}{Method} & \multicolumn{4}{c|}{Target domain} & \multirow{2}{*}{Avg.($\uparrow$)}\\
    \cmidrule(r){2-5}
     & L100 & L38 & L43 & L46 \\
    \midrule
    VREx \cite{krueger2021out} (ICML'2021)   & $40.65$         & $29.95$        & $50.06$         & 33.72          & $38.60$\\
    MTL \cite{blanchard2021domain} (JMLR'2021)    & $38.94$         & $35.18$        & $52.80$         & $35.29$        & $40.55$ \\
    SagNet \cite{nam2021reducing} (CVPR'2021)  & 47.25           & 29.67          & 52.87           & 25.22          & $38.75$\\
    ARM \cite{zhang2021adaptive} (NeruIPS'2021)   & $44.98$         & $33.73$        & $43.39$         & $27.77$        & $37.47$\\
    SAM \cite{foret2021sharpnessaware} (ICLR'2021)   & 55.66         & 27.92        & 51.51  & 31.93       & 41.76\\
    FACT \cite{xu2021fourier} (CVPR'2021)   & 52.90        & 38.66        & 52.32         & 31.58       & 43.87\\
    SWAD \cite{cha2021swad}  (NeurIPS'2021)   & $49.80$         & $33.16$        & \bfseries55.57  & $33.19$        & $42.93$\\
    MIRO \cite{cha2022domain} (ECCV'2022) & 53.78           & 31.88          & 51.67           & 33.21          & 42.63\\
    PCL \cite{yao2022pcl}  (CVPR'2022)   & $52.62$         & 39.98        & $48.49$         & $31.74$        & $43.21$\\
    AdaNPC \cite{zhang2023adanpc} (ICML'2023)  & 50.60           & 38.60         & 42.20         & 34.00        & 41.35\\
    DandelionNet \cite{hu2023dandelionnet}  (ICCV'2023) & 52.78           & 32.80         & 50.69         & 31.63        & 41.98\\
    iDAG \cite{huang2023idag}  (ICCV'2023) & 53.78           & 34.82         & 50.28         & 28.85        & 41.93\\
       SAGM \cite{wang2023sharpness} (CVPR'2023) & 50.20 & 27.54 & 53.21 & 31.70 & 40.66 \\
       GMDG \cite{tan2024rethinking} (CVPR'2024)& 50.70  & 34.78 & 51.26 & 36.63 & 43.34\\
    \midrule
    {DDG (ours)}& \bfseries56.10          & \bfseries42.67          & 51.03         & \bfseries36.82         & \bfseries46.63 \\
    \bottomrule
    \end{tabular}
    \label{terra}
    \vspace{-3mm}
\end{table}

\begin{table}[t]
    \centering
    \setlength{\tabcolsep}{1.5pt}
    \caption{Generalization performance on VLCS with object recognition accuracy~(\%) based on ResNet-18 pre-trained on ImageNet. High is better, and bold indicates the best performance.}
    {
    \begin{tabular}{c|cccc|c}
    \toprule
    \multirow{2}{*}{Method} & \multicolumn{4}{c|}{Target domain} & \multirow{2}{*}{Avg.($\uparrow$)}\\
    \cmidrule(r){2-5}
     & Caltech & LabelMe & SUN & PASCAL \\
    \midrule
    VREx \cite{krueger2021out} (ICML'2021)   & $96.20$         & $62.97$        & 73.65  & 73.68          &  $76.62$\\
    MTL \cite{blanchard2021domain} (JMLR'2021)     & $96.38$         & $62.54$        & $70.91$         & $71.68$        & $75.38$ \\
    SagNet \cite{nam2021reducing} (CVPR'2021)  & 97.09           & $62.07$        & $70.37$         & 75.42          & 76.24\\
    ARM \cite{zhang2021adaptive} (NeurIPS'2021)    & $96.29$         & $61.55$        & $72.32$         & $76.27$        & $76.61$\\
    SAM \cite{foret2021sharpnessaware}  (ICLR'2021)       & 98.15         & 60.52         & 71.25          & 75.90 & 76.45\\
    FACT \cite{xu2021fourier}  (CVPR'2021)        & 97.10           & 63.25          & 72.67           & 74.97       & 77.00 \\
    SWAD \cite{cha2021swad}  (NeurIPS'2021)        & 97.70           & 61.27          & 70.72           & 76.71 & 76.60\\
   MIRO \cite{cha2022domain} (ECCV'2022)        & 97.79           & 61.98         & 71.21           & 74.53          & 76.38\\
    PCL \cite{yao2022pcl}  (CVPR'2022)         & 97.09           & 62.07          & 71.06           & 75.05          & 76.32\\
     AdaNPC \cite{zhang2023adanpc}  (ICML'2023) & 98.00           & 60.20        & 69.10         & 76.60        & 75.98\\
     DandelionNet \cite{hu2023dandelionnet}  (ICCV'2023) & 94.61           & 63.06        & 67.17        & 71.49       & 74.08\\
     iDAG \cite{huang2023idag}  (ICCV'2023)  & 94.44           & 59.88        & 70.18         & 72.86        & 74.34\\
       SAGM \cite{wang2023sharpness} (CVPR'2023) & 96.03 & 60.99 & 70.64 & 75.68 & 75.83 \\
       GMDG \cite{tan2024rethinking} (CVPR'2024)& 96.56  & 63.53 & 69.35 & 73.83 & 75.81\\
    \midrule
    {DDG (ours)}       & \bfseries99.08           & \bfseries63.69          & \bfseries73.87           & \bfseries76.78          & \bfseries78.36 \\
    \bottomrule
    \end{tabular}} 
    \vspace{-2mm}
    \vspace{-3mm}
    \label{VLCS}
\end{table}
\noindent\textbf{Results on PACS.} TABLE~\ref{PACS} reports the generalization performance on PACS, demonstrating the superior generalization performance achieved by our proposed DDG. Specifically, our DDG outperforms the state-of-the-art method iDAG by 1.5\% in terms of the model's average generalizability. Notably, on hard-to-transfer domains where the distribution variance is substantial and existing methods exhibit poor performance, such as `Sketch' within PACS, DDG markedly improves generalization performance by 9.6\% compared to the state-of-the-art method iDAG, signifying its efficacy in preventing overfitting in tasks~(\emph{e.g.}, `Photo') that are already near saturation in performance, while preserving crucial semantic features. These findings underscore the superiority of our proposed DDG in capturing genuine semantic information rather than pixel-level details.

\noindent\textbf{Results on TerraIncognita.} We conclude the results on TerraIncognita in TABLE~\ref{terra}. As observed, our DDG emerges with the top performance in three out of the four scenarios, with a substantial improvement of 6.2\%, 10.4\%, and 16.6\% on L100, L38, and L46, respectively, compared to the state-of-the-art method FACT. Besides, our DDG obtains a performance gain of 6.4\% over FACT in terms of average generalization performance, highlighting the efficacy of our DDG in tackling distribution shifts across domains.

\noindent\textbf{Results on VLCS.} The generalization performance on VLCS is summarized in TABLE~\ref{VLCS}. Notably, the proposed DDG achieves the highest generalization performance across all the scenarios, with improvements of 2.0\%, 1.3\%, 3.3\%, and 2.4\% on Caltech, LableMe, SUN, and PASCAL, respectively. As a result, our DDG maintains an enhancement of 2.3\% compared to the state-of-the-art approach in average generalization performance. These outcomes collectively underscore that our proposed DDG, employing vector quantization, enhances generalization performance by prioritizing semantic information over pixel-level details.

\end{document}